\documentclass[10pt,conference]{IEEEtran}
\usepackage{cite}
\usepackage[dvips]{color}
\usepackage{tabu}
\usepackage{comment}
\usepackage{todonotes}
\usepackage{epsf}
\usepackage{epsfig}
\usepackage{times}
\usepackage{epsfig}
\usepackage{graphicx}

\usepackage{mathtools}
\usepackage{mathrsfs}
\usepackage{amssymb,amsmath}
\usepackage{amsmath}
\usepackage{amsfonts}
\usepackage{booktabs} 
\usepackage{multirow} 
\usepackage{array} 

\usepackage{algorithmicx}  
\usepackage[algo2e]{algorithm2e} 
\usepackage{cleveref}
\usepackage{dsfont}
\usepackage{lettrine} 
\usepackage{epsfig,algorithm,algpseudocode,amsthm,url}
\usepackage{caption}
\usepackage{subcaption}
\usepackage[justification=raggedright]{caption}
\usepackage{subcaption}
\usepackage{bbm}
\usepackage[utf8]{inputenc}
\usepackage{verbatim}
\usepackage{setspace}	
\usepackage[T1]{fontenc}
\usepackage{textcomp}
\usepackage{tabularx}
\usepackage{xpatch}

\usepackage{stackengine}
\usepackage[nocomma]{optidef}
\usepackage{commath}

\usepackage[left=1.62cm,right=1.62cm,top=1.85cm,bottom=2.6cm]{geometry}

\IEEEoverridecommandlockouts 

\renewcommand{\baselinestretch}{0.94}  

\begin{document}
\title{\Huge Thermal Image Calibration and Correction using Unpaired Cycle-Consistent Adversarial Networks \thanks{This material is based upon work supported by the Air Force Office of Scientific Research under award number FA9550-20-1-0090, the National Aeronautics and Space Administration (NASA) under award number 80NSSC23K1393, and the National Science Foundation under Grant Numbers CNS-2232048, and CNS-2204445.}

}
\author{
	\IEEEauthorblockN{
	Hossein Rajoli, Pouya Afshin, Fatemeh Afghah}

	\IEEEauthorblockA{Holcombe Department of Electrical and Computer Engineering, Clemson University, Clemson, SC, USA \\
Emails: hrajoli@clemson.edu, pafshin@clemson.edu, fafghah@clemson.edu}
}
\maketitle\vspace{-0.5cm}
\begin{abstract}
Unmanned aerial vehicles (UAVs) offer a flexible and cost-effective solution for wildfire monitoring. However, their widespread deployment during wildfires has been hindered by a lack of operational guidelines and concerns about potential interference with aircraft systems. Consequently, the progress in developing deep-learning models for wildfire detection and characterization using aerial images is constrained by the limited availability, size, and quality of existing datasets. This paper introduces a solution aimed at enhancing the quality of current aerial wildfire datasets to align with advancements in camera technology. The proposed approach offers a solution to create a comprehensive, standardized large-scale image dataset.
This paper presents a pipeline based on CycleGAN to enhance wildfire datasets and a novel fusion method that integrates paired RGB images as attribute conditioning in the generators of both directions, improving the accuracy of the generated images. 
\end{abstract}
\section{Introduction} \vspace{-0cm}

In recent years, due to climate changes, wildfire has emerged as a major area of research, drawing the attention of scholars worldwide. Researchers are focused on gaining a better understanding of the different types of wildfires, predicting their behavior, and incorporating fire-physics information to develop more effective strategies for fire management \cite{jain2020review}. This
increasing emphasis on wildfires is driven by their devastating impact on the natural environment and the global economy, resulting in significant financial losses and extensive damage to homes, businesses, wildlife, and the environment \cite{vera2023wildfire}. Advanced technologies such as drones improve fire management by offering swift deployment and adaptable imaging capabilities. Recent drone-based research focuses on solving various fire management challenges \cite{afghah2019wildfire,chen2022wildland, Shamsoshoara_DDDAS2023,9217742,boroujeni2024ic}.

AI has become an indispensable technology in our daily lives, with numerous applications across various fields \cite{sam2023stability,behjoo2023exact,samanipour2023, kargar2023integrated}. Within the realm of wildfire management, Deep Learning (DL) methods have taken precedence in existing aerial monitoring systems, providing a promising path for efficient wildfire assessment \cite{tang2021attentiongan}. Leveraging DL techniques, these systems enable accurate flame detection \cite{bouguettaya2022review}, smoke characterization \cite{sathishkumar2023forest}, and prediction of fire spread \cite{mccarthy2021deep}, marking a significant advancement in wildfire monitoring capabilities.
However, while these systems rely heavily on visual processing through cameras, limitations in modern high-quality Infra-Red (IR) cameras pose challenges due to their high costs, distance restrictions, and lack of color information \cite{sun2022drone}.

Obtaining comprehensive datasets is vital for the effectiveness of DL-based methods. However, the collection of drone-based fire data poses challenges in terms of cost and logistics. Existing datasets, collected by different cameras \cite{fule2022flame, Shamsoshoara_DDDAS2023}, may exhibit variations in image quality that do not align with desired standards. This study addresses this issue by developing a pipeline that enhances existing datasets, aiming to create a unified large-scale image dataset that meets the benchmarks set by newer datasets collected with advanced cameras, known for their reduced artifacts and minimized drone wobbling. This enhancement facilitates streamlined data integration, fostering the creation of cohesive and consolidated datasets crucial for advancing DL-based methods. The proposed approach serves as a practical solution to overcome challenges associated with aerial data collection.

Innovative contributions in 
image-to-image translation, specifically in Generative Adversarial Network (GAN) frameworks, presents potential directions for 
this purpose \cite{tang2021attentiongan, goodfellow2020generative}. These techniques offer a potential path to improve wildfire monitoring and temperature analysis by enhancing existing datasets. While previous studies focused on fusion-related research for infrared images, specifically thermal images \cite{ciprian2021fire}, none have explicitly aimed to bridge the quality and standard-related gap between new high-tech images and existing datasets by upgrading them. 
In this paper, we proposed a fire image calibration model based on CycleGAN noting its promising performance in different image-to-image translation tasks, its robustness to domain shifts in the data, and its ability to perform image translation without the need for paired training data.


The contributions of this work are summarized as follows:

\begin{itemize}
    \item A  non-symmetric paired generator based on CycleGAN 
    is developed for IR fire image calibration
    that can be used to enhance wildfire datasets, bringing them up to the standards of high-tech quality tools.

    \item 
    A novel multi-level architecture is proposed for generators, which enhances qualitative performance by injecting local features from various-level receptive fields into the global features pool.

    \item A novel upsampling method is offered that can preserve the target domain resolution utilizing flexible convolution.
\end{itemize}\vspace{-0cm}

The subsequent sections of this paper are organized as follows. Section 2 provides an overview of the existing literature concerning image translation. Section 3 details the proposed framework, encompassing the network architecture, loss functions, and the training process. Section 4 presents and analyzes the experimental outcomes.

\section{Related Works}\label{relworks}
Within the fire detection domain, GAN models have primarily been used for tasks such as converting data modalities \cite{boroujeni2024ic}, compositing flame or smoke images to augment datasets for improved training \cite{boone2023attention}, and generating synthetic datasets \cite{carreon2023generative}. To the best of our knowledge, this is the first attempt to leverage GAN models explicitly for dataset matching within the 
wildfire management domain, aiming to address the limitations in generating comprehensive and large-scale fire spread images.
Recent advancements in GAN models have introduced various architectures for diverse image translation tasks. These models offer distinct methodologies in image translation. In continuing, we discuss their applicability, strengths, and limitations concerning the proposed image translation problem.

Inspired by natural language translation techniques, \cite{yi2017dualgan} proposed Dual-GAN that utilizes two generators to perform image-to-image translation effectively. It aims for universal solutions in image translation without domain-specific knowledge. Later, a versatile generative network, namely Star-GAN proposed by \cite{choi2018stargan}, achieves high-quality translations across multiple domains, offering flexibility in image manipulation during testing. However, Star-GAN suffers from limitations in output control due to multi-domain training and may lose high-resolution details, especially in complex tasks, despite its multi-task learning capabilities \cite{choi2018stargan}.
Moving beyond general image-to-image translation methods, SAR-CGAN emerges as a Conditional GAN specifically tailored for Synthetic Aperture Radar (SAR) image generation \cite{yang2023unpaired}. It introduces improvements in loss functions to enhance image accuracy and detail features, offering advantages in generating SAR images resembling ground truth data and handling high-quality outputs from low-quality inputs.
such as the need for ample training samples, inability to capture real-world complexities, and the potential need for post-processing still persist \cite{yang2023unpaired}.
Incorporating a ResNet Pix2Pix translation network and spectral indices module, \cite{hu2023gan} introduced SARO-GAN, aiming for stable training and diverse image generation. Despite demonstrating stability and diversity, challenges include dataset heterogeneity, capturing fine details in some optical images, and limited evaluation metrics.

In \cite{boone2023attention}, an attention-guided GAN-based image-to-image translation tool generates wildfire images from aerial forestry images, aiming to improve classification accuracy. Highlighting the significance of attention mechanisms in generating wildfire images via image-to-image translation techniques, their study showcases increased accuracy in classifying a distinct test set from the training dataset when supplemented with diverse synthetic wildfire images. Conversely, \cite{boroujeni2024ic} introduces a novel DL architecture centered on deriving temperature information from aerial true-color RGB images by transforming them into the Infrared Radiation (IR) domain. The authors proposed an architecture, namely the Improved Conditional Generative Adversarial Network (IC-GAN), that employs matched IR images as a guiding condition for the generator during the translation process. Their approach involves a U-Net-based generator coupled with a mapper module to convert outputs into various color spaces with adaptable parameters. To ensure structural similarity without unwarranted pixel-level disparities, \cite{boroujeni2024ic} integrates clustering alignment into the loss function.

We address all shortcomings of the reviewed papers by introducing multi-level flexible convolution-based non-symmetric generators that cooperate in a cycle-GAN framework to translate a low-quality domain to a high-quality standard temperature-aware IR domain.

\section{Proposed Method}\label{sysmodl}
In this part, we delve into the proposed architecture for transforming two infrared (IR) image domains, specifically tailored for UAV monitoring systems. In three sections we elaborate on our method. Section \ref{proposArch} explains how the two generators are structured in a non-symmetric cycle-GAN system. Next, Section \ref{PropDiscArch} delves into the specific loss functions used for these networks. Lastly, Section \ref{lossFunctions} outlines the loss functions utilized in the training procedure of our method.

\subsection{Proposed Architecture} \label{proposArch}

\begin{figure*}[t] 
    \centering
    \includegraphics[width=\linewidth]{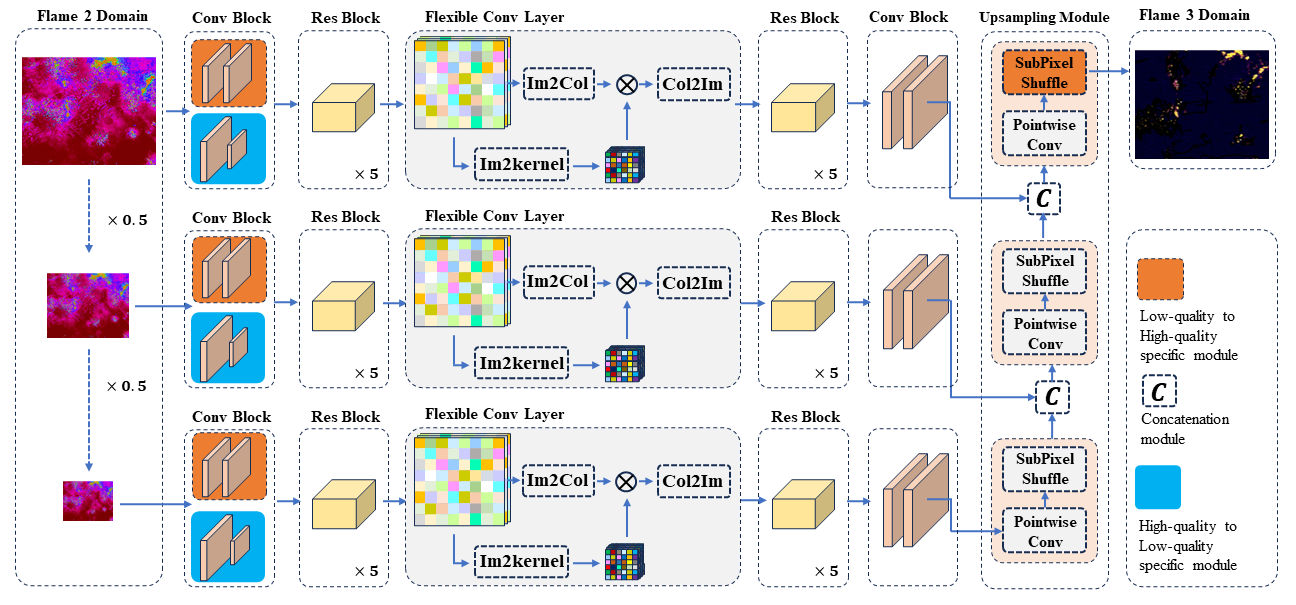}
    \caption{The schematic diagram of both proposed generators}
    \label{fig:genLow2High}
\end{figure*}

CycleGAN is a GAN that can translate images between two different domains without using paired training data. It employs two generators and discriminators to learn mappings between the domains and maintain cycle consistency constraints to keep the generated images close to the input domain.
Based on \cite{boone2023attention}, existing cycle-GAN based methods struggle with diverse forestry IR scenes when generating synthetic IR images, often causing unwanted artifacts or drastic alterations in input images. This underscores the need for an AttentionGAN specifically designed for generating synthetic wildfire images from varied inputs. Another promising approach involves flexible convolution mechanisms that allocate more computational resources to capture local features, akin to pixel-wise attention on the feature map level. These mechanisms address key challenges in domain translation: firstly, they've been successful in denoising scenarios, crucial for achieving high-quality IR translations; secondly, their resemblance to pixel-shuffling techniques common in super-resolution tasks aids in addressing super-resolution issues in high-quality domain translation. Moreover, the framework's capability to capture and inject local features into the global context is crucial, given the distributed and less distinguishable nature of IR domain features, particularly heating features, unlike the more separable objects in the RGB domain. The multi-level flexible convolution non-symmetric generators, as depicted in Fig.~\ref{fig:genLow2High}, inherently possess these three critical features. \vspace{-0.2cm}

\subsection{Non-Symmetric Generator Architecture} \label{PropGenArch}

Our pipeline includes two non-equal generator modules, $G_{AB}$ and $G_{BA}$. The proposed generator architecture incorporates a multi-tier feature extraction framework, comprising three distinct levels of input, each downsampled by a factor of 0.5 in both width (W) and height (H) dimensions. Following this downsampling, the inputs undergo sequential convolutional operations and flexible convolution layers, facilitating adaptive local feature capture within the network.

Subsequently, the outputs from these layers undergo an upsampling process using pixel shuffling, a method designed for effective upsampling. Pixel shuffling reorganizes feature maps to augment resolution, reconstructing higher-fidelity images. These upsampled outputs from the three hierarchical levels are then concatenated, potentially utilizing concatenation layers, to form the final output of the generator network. This concatenation, representing the fusion of features across multiple scales and hierarchical levels, signifies the injection of local features captured from varying-sized receptive fields into the higher-level layer's wider receptive field features. This process aids the network in integrating information gathered from diverse scales, enriching its understanding of complex spatial patterns and enhancing the depth and diversity of the generated outputs. 

This architecture leverages multiple advantages. Firstly, the multi-level feature extraction allows the network to extract hierarchical representations at various levels of abstraction, enabling the processing of information across multiple scales. The integration of downsampling and upsampling stages facilitates comprehensive information processing, capturing both fine-grained and high-level features. Additionally, pixel shuffling during upsampling aids in preserving spatial information, crucial for reconstructing higher-resolution outputs with improved fidelity. Finally, the concatenation of features extracted from different hierarchical levels enables the network to fuse diverse information, potentially enriching the complexity and depth of the generated outputs. As an input to generators, we use a fusion module to incorporate the corresponding RGB images as attribute conditioning for the generators, allowing them to learn more effectively from both datasets, however, during the test phase we disable the module by feeding a zero-value image to it.

\subsection{Discriminator Architecture} \label{PropDiscArch}

In our framework, inspired by the discriminator proposed by \cite{boroujeni2024ic} we employ a Patch-GAN strategy, segmenting input images into non-overlapping patches for individualized processing. By analyzing each patch separately through CNNs, it comprehensively evaluates the likelihood of authenticity for the generated images. This approach enables detailed assessment, capturing local nuances and textures while preserving overarching global features. \vspace{-0.4cm}

\begin{figure}
    \centering
    \includegraphics[width = \linewidth ]{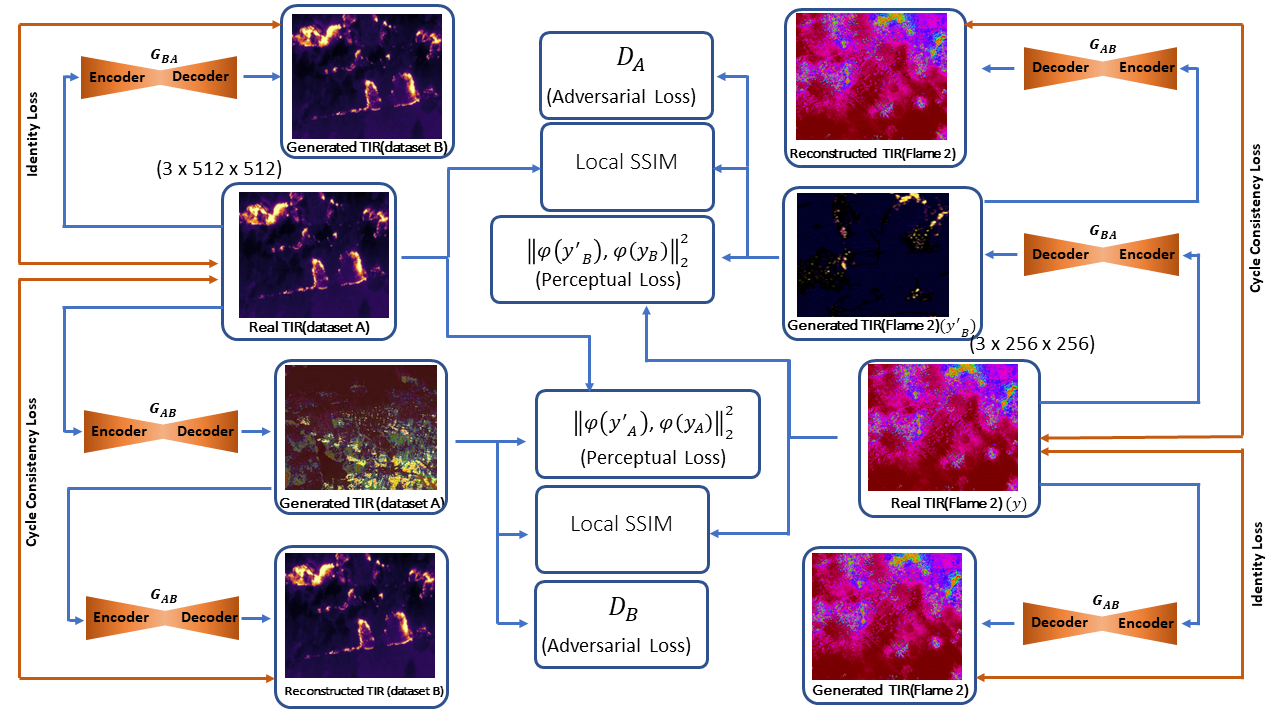}
    \caption{The schematic diagram of the proposed pipeline}\vspace{-0.2cm}
    \label{fig:pipeline}\vspace{-0.1cm}
\end{figure}

\subsection{Loss Functions} \label{lossFunctions}

The loss function for the generators is a combination of several terms.\\
\noindent \textbf{Adversarial loss} which encourages $G_{AB}$ to generate images that are indistinguishable from the real low-quality thermal images. $L_{GAN}(G_{AB}, D_{B}, A, B) = \mathbb{E}{b \sim p_{data}(B)}[\log D_{B}(b)] + \mathbb{E}{a \sim p_{data}(A)}[\log(1 - D_{B}(G_{AB}(a)))]$, where $D_B$ is the discriminator for images in set $B$, $G_{AB}$ is the generator that maps domain $A$ to $B$. Samples selected from domain $A$ and $B$ are $a$ and $b$ respectively.

\begin{figure}
    \centering
    \includegraphics[width = \linewidth ]{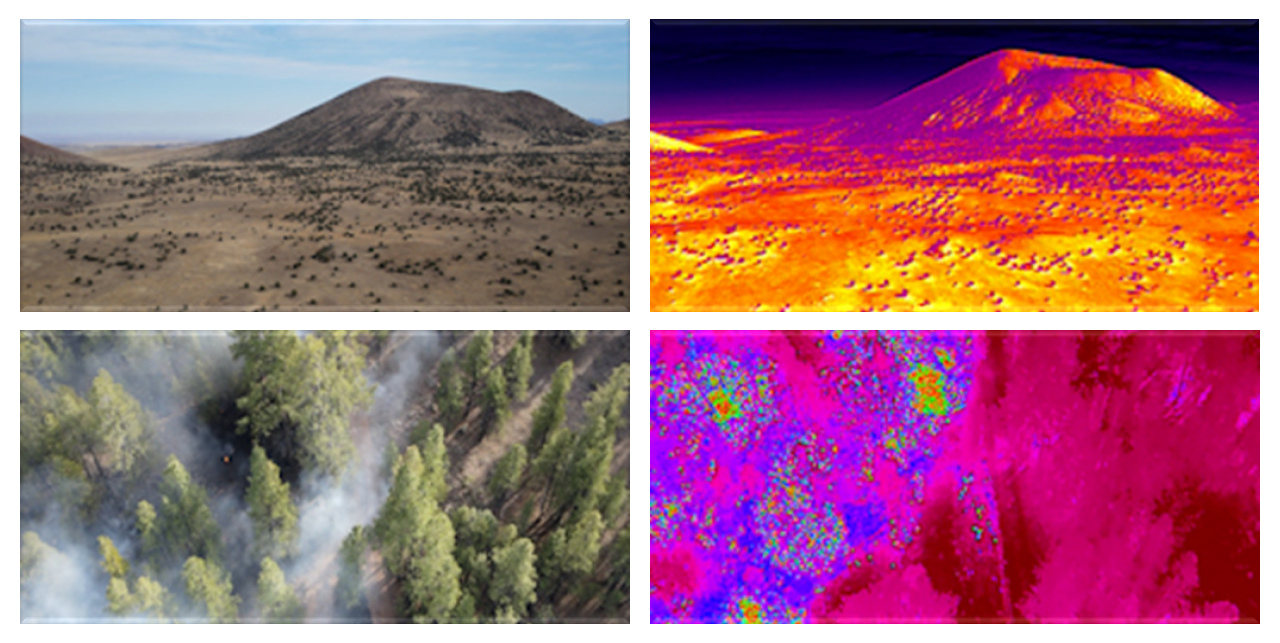}
    \caption{\small{Flame 2 samples. Due to lack of temperature reference, cold areas are demonstrated with red color.}}\vspace{-0.2cm}
    \label{flame2_samples}\vspace{-0.1cm}
\end{figure}

\noindent
\textbf{Cycle-consistency loss} which enforces the cycle-consistency property, i.e., the mapping of an image from set $A$ to set $B$ and then back to set $A$ should result in the original image, and vice versa. $L_{cyc}(G_{AB}, G_{BA}, A, B) = \mathbb{E}{a \sim p_{data}(A)}[||G_{BA}(G_{AB}(a)) - a||_1] + \mathbb{E}{b \sim p_{data}(B)}[||G_{AB}(G_{BA}(b)) - b||_1]$. 

\noindent \textbf{Identity loss} which encourages $G_{AB}$ to preserve the content of low-quality thermal images and $G_{BA}$ to preserve the content of high-quality thermal images. $L_{id}(G_{AB}, G_{BA}, A, B) = \mathbb{E}{b \sim p_{data}(B)}[||G_{AB}(b) - b||_1] + \mathbb{E}{a \sim p_{data}(A)}[||G_{BA}(a) - a||_1]$. It means if an image from domain $A$ is fed into $G_{BA}$ it should result in itself. 

\noindent \textbf{Structural Similarity Index Measure (SSIM):} The SSIM-based loss is a vital metric used to assess the resemblance between images based on their structural characteristics, encompassing edges and textures \cite{borji2019pros}. 
\begin{equation}
L_{SSIM(x, y)} = 1- \frac{{(2\mu_x\mu_y + c_1)(2\sigma_{xy} + c_2)}}{{(\mu_x^2 + \mu_y^2 + c_1)(\sigma_x^2 + \sigma_y^2 + c_2)}}
\end{equation}
The means, $\mu_x$ and $\mu_y$, variances, $\sigma_x$ and $\sigma_y$, and covariance, $\sigma_{xy}$, of images assess the average intensity and spread. These metrics serve to evaluate the structural similarity of images: high values show a more pronounced resemblance in features such as edges and textures, whereas lower scores highlight substantial differences.
Overall, a high SSIM score signifies a striking similarity between the compared images, indicating minimal discrepancies in their structural information. The formula for SSIM encapsulates these structural attributes and calculates the degree of similarity between two images. It is a fundamental measure utilized in evaluating image fidelity and resemblance within various domains. In our work, we leverage SSIM-based loss two times;
\begin{itemize}
    \item $L_{DSSIM}$: it computes SSIM-based loss between extracted high-semantic features by means of network $\phi$ in perceptual loss.
    \item  $L_{SSIM}$: it calculates SSIM-based loss on low-semantic features of the original and generated image.
\end{itemize}
\textbf{Perceptual loss:} Inspired by \cite{yang2023unpaired}, to alleviate distortion caused by the adversarial loss, a new term of loss, namely perceptual loss, will be added to the loss function as follow: $L_{perc}(G_{AB}, G_{BA}) = \frac{1}{N}\sum_{i=1}^{N}\left\lVert \phi(I_{A, i}) - \phi(G_{BA}(I_{B, i})) \right\rVert_1 + \frac{1}{N}\sum_{i=1}^{N}\left\lVert \phi(I_{B, i}) - \phi(G_{AB}(I_{A, i})) \right\rVert_1 + \lambda L_{DSSIM}$, where $\phi$ is a pre-trained network of ResNet-18, and $N$ is the number of samples in the dataset. $L_{DSSIM}$ here refers to SSIM-based loss applied to the deep features extracted by $\phi$.

\vspace{-0.2cm} 
\section{Evaluation}
 
In this section, we evaluate the performance of our CycleGAN-based model on wildfire imagery. 
\vspace{-0.2cm}
\subsection{Dataset}
Most of the currently available wildfire presenting paired RGB-IR images same as FLAME 2 lacks images calibrated to the temperature reference. This flaw results in presenting the hottest detectable spot in the image as red in Fig.~\ref{flame2_samples}, irrespective of its actual temperature. This inherent limitation within the dataset introduces ambiguity during model training, posing a significant challenge. These challenges are what we aim to address by mapping samples to a more precise representation domain.
Our evaluation employs two different datasets for testing the performance of our CycleGAN model. Firstly, we use the Flame II dataset \cite{fule2022flame}, which consists of 53,451 paired RGB and thermal infrared (TIR) aerial images captured during a prescribed fire in Northern Arizona in November 2021. These images were recorded in video format, offering a rich source of data for our evaluation. Additionally, we acquired a new dataset from the Cycan Marsh wildfire, as part of the Flame III dataset, using a DJI Mavic 2 Enterprise Advanced drone equipped with a thermal camera with a resolution of $640 \times 512$ and a frame rate of $30 Hz$, providing a high-tech thermal reference for our experiments. These datasets provide a diverse range of environmental conditions and image qualities, enabling us to evaluate the robustness and generalization capabilities of our model. We report our findings on the effectiveness of our approach using these datasets in the following subsections.
\subsection{Implementation and Results}
We employed an NVIDIA 3080 GPU for rapid training of the CycleGAN-based thermal image enhancement model, optimizing parameters like batch size (32) and learning rate (0.0002). Implemented in Pytorch, our model enhances generated thermal images by leveraging information from both RGB and thermal modalities. This model implements early concatenation fusion and a pre-trained Resnet-18 for calculating perceptual loss, leveraging insights from a wildfire dataset.\vspace{-0cm}

\begin{figure}[t]
    \centering
    \includegraphics[width = \linewidth ]{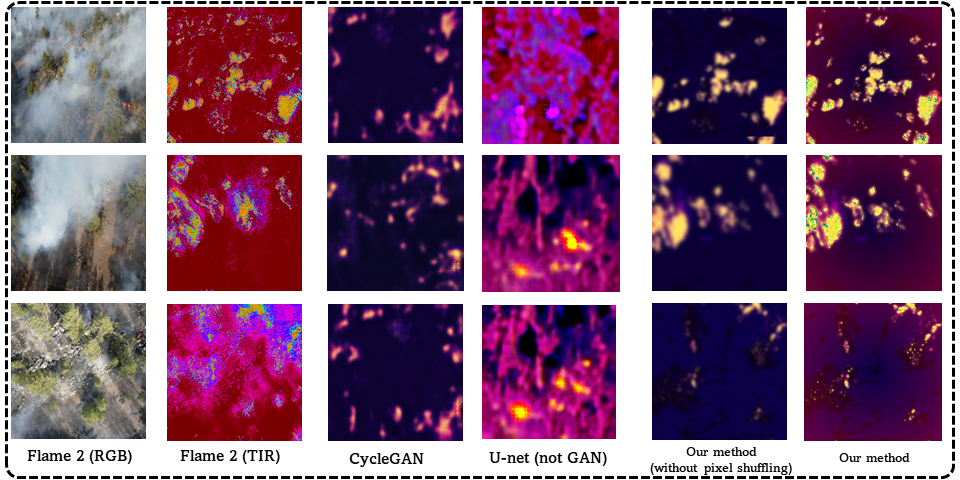}
    \caption{\small{Qualitative results from 50 epochs comparing the proposed IR-to-IR image translation method against other translation models. The first two Columns depict the input RGB and thermal images, followed by outputs from Cycle-GAN, U-Net, and our method, both with and without the super-resolution upsampling module.}}\vspace{-0.2cm}
    \label{fig:qualitative Comp.}\vspace{-0.1cm}
\end{figure}

The examination of generative models, employing both perceptual loss and SSIM metrics, illuminates distinct performance nuances among the methods. Compared to the baseline, our Model (ours + $L_\phi$) — proposed model fortified solely with tailored perceptual loss—reveals a significant reduction in average perceptual loss, indicating improved image reconstruction fidelity.
Introducing SSIM loss in our Model (ours + $L_\phi$ + $L_{SSIM}$) demonstrates a marked increase in the SSIM index when compared to the other models. The integration of both customized perceptual loss and SSIM loss in our proposed model notably advances the structural similarity between the generated and target images.
These observed enhancements across metrics validate the effectiveness of our proposed model. The model equipped with tailored perceptual loss and SSIM loss components, emerges as a promising approach, exhibiting superior fidelity and structural similarity in synthesized images compared to the baseline.
Fig.~\ref{fig:qualitative Comp.} presents our proposed framework's qualitative comparison with other methods, demonstrating superior image resolution, diversity, and alignment accuracy. Our method consistently delivers high-quality images across various properties, unlike U-Net or cycle-GAN which struggle to reconstruct IR image details. Although cycle-GAN shows improvement by capturing the overall heat map, it fails to pinpoint flame positions accurately and generates irrelevant spots in the IR domain, indicating a deficiency in domain feature preservation and alignment.
Surprisingly, even without high-resolution pixel-shuffling upsampling, our proposed model outperforms competitors, providing structurally accurate and detailed IR images, albeit with a lack of high resolution and slight blurriness. Notably, our full-stack method excels at identifying small fire zones, depicting them with green or reddish shades while preserving structural similarity between translated and reference IR images, regardless of background complexity. \vspace{-0.4 cm}

\begin{table}[h]
\centering
\caption{\small{ Average $L_{\phi}$ and $\text{SSIM}$ over the test set. Here $\text{SSIM}$ is calculated on high-resolution images and $L_{\phi}$ is a pre-trained VGG\_19.}}
\vspace{0pt}
\label{tab:Performance Metrics}

\renewcommand{\arraystretch}{.05}
\begin{tabular}
{ >{\centering\arraybackslash}p{3.5cm}
 >{\centering\arraybackslash}p{2cm}
 >{\centering\arraybackslash}p{2cm}
  }
\toprule

{\fontsize{9}{8}\selectfont { \rule{-5pt}{0ex}
\begin{center}
\textbf{Model}
\end{center}
}}
&
{\fontsize{9}{8}\selectfont { \rule{-5pt}{0ex}
\begin{center}
\textbf{Average $L_{\theta}$}
\end{center}
}}
&
{\fontsize{9}{8}\selectfont { \rule{-8pt}{0ex}
\begin{center}
\textbf{Average SSIM}
\end{center}
}}
\\[-.5ex]
 \midrule
 
{\fontsize{8}{7}\selectfont { \rule{-3pt}{0ex}
Baseline (Cycle-GAN)
}}
&
{\fontsize{8}{7}\selectfont {\rule{-5pt}{0ex}
0.67
}}
&
{\fontsize{8}{7}\selectfont {\rule{-5pt}{0ex}
0.31
}}

\\[-1ex]

{\fontsize{8}{7}\selectfont {\rule{-3pt}{0ex}
Ours ($L_\phi$)
}}
&
{\fontsize{8}{7}\selectfont {\rule{-5pt}{0ex}
0.41
}}
&

{\fontsize{8}{7}\selectfont {\rule{-5pt}{0ex}
0.39
}}
 
\\[-1ex]
{\fontsize{8}{7}\selectfont {\rule{-3pt}{0ex}
Ours ($L_{\phi} + L_{SSIM}$)
}}
&
{\fontsize{8}{7}\selectfont {\rule{-5pt}{0ex}
0.33
}}
 &

{\fontsize{8}{7}\selectfont {\rule{-5pt}{0ex}
0.54
}}

\\[-2ex]
\bottomrule
\end{tabular}
\end{table}

\vspace{-0.5cm}
\section{Conclusion}\label{conclusion}
\vspace{-0.1cm}
The objective of this paper is to augment the existing fire IR image dataset by mapping samples to an IR domain with a temperature reference, increasing their resolution, and reducing noise and artifacts.
For this purpose, we proposed a model that combines the CycleGAN idea and a fusion network that utilizes RGB information to effectively enhance the quality of thermal images. The method has shown potential in improving images obtained from less advanced technologies to match the quality of those acquired by high-tech tools. As a result, this approach could find applications in various fields where thermal imaging is utilized.\vspace{-0.3cm}

\def\baselinestretch{0.91}
\bibliographystyle{IEEEbib}
\bibliography{Main}
\end{document}